# An Experimental Comparison of Numerical and Qualitative Probabilistic Reasoning

**Max Henrion**     **Gregory Provan**
Institute for Decision Systems Research
4894 El Camino Real, Suite 110, Los Altos, CA 94022
henrion / provan@camis.stanford.edu

Brendan Del Favero[1]     Gillian Sanders[2]
[1]Department of Engineering-Economic Systems,
Stanford University, CA 94305
[2]Section on Medical Informatics
Stanford University, CA 94305

## Abstract

Qualitative and infinitesimal probability schemes are consistent with the axioms of probability theory, but avoid the need for precise numerical probabilities. Using qualitative probabilities could substantially reduce the effort for knowledge engineering and improve the robustness of results. We examine experimentally how well infinitesimal probabilities (the kappa-calculus of Goldszmidt and Pearl) perform a diagnostic task — troubleshooting a car that will not start — by comparison with a conventional numerical belief network. We found the infinitesimal scheme to be as good as the numerical scheme in identifying the true fault. The performance of the infinitesimal scheme worsens significantly for prior fault probabilities greater than 0.03. These results suggest that infinitesimal probability methods may be of substantial practical value for machine diagnosis with small prior fault probabilities.

## 1 BACKGROUND AND GOALS

Bayesian and decision theoretic methods have long been criticized for an excessive need for quantification. They require many numerical probabilities and utilities that are difficult to assess and are liable to judgmental biases. Some people claim that since human thinking is inherently qualitative, it is incompatible with quantitative schemes. These criticisms have fueled interest in alternative formalisms for reasoning and decision making under uncertainty that are intended to be easier to use and more compatible with human cognition. Among these alternative schemes are: various generalizations of decision theory [Edwards, 1992]; Dempster-Shafer belief functions [Shafer, 1976]; generalizations of logic, including default and non-monotonic logics [Ginsberg, 1987]; fuzzy logic [Zadeh, 1983]; possibility theory [Dubois and Prade, 1988]; and fuzzy probabilities.

If, however, our goal is simply to provide a qualitative basis for reasoning and decision making under uncertainty, there is no need to abandon Bayesian decision theory. The axioms of decision theory, indeed, assume only the ability to make qualitative judgments – that is, to order events by probability or outcomes by desirability. The quantification of probabilities and utilities can be based on purely qualitative judgments. Furthermore, several schemes have been developed that are purely qualitative, but are consistent with the axioms of decision theory.

One such scheme is *qualitative probabilities*, originated by Wellman [1990; Henrion & Druzdzel 1991; Wellman & Henrion, 1993]. A second approach to qualitative probabilities is the kappa-calculus [Goldszmidt and Pearl, 1992], which represents all probabilities in a Bayesian belief network by $\varepsilon^\kappa$, where $\kappa$ is an integral power of $\varepsilon$. The $\kappa$-calculus is consistent with the axioms of probability where ε→0. Events are ranked according to κ. Events with larger κ are assumed to be negligible relative to events with smaller κ. The calculus provides a plausible set of events: those with the smallest (most probable) consistent with the observed findings. The calculus is sometimes called *qualitative probability*. To avoid confusion with other qualitative probability schemes, we call this representation *infinitesimal probabilities*. Pearl [1993] has extended this scheme to handle qualitative utilities to support decision making.

The κ–calculus or infinitesimal probabilities can be looked at in two ways: (a) as providing a scheme for non-monotonic reasoning whose semantics are firmly grounded in probability and decision theory; or (b) as providing a simplification of belief networks with numerical probabilities. In this paper, we are focus on the second view, and examine the performance of infinitesimal probabilities as an approximation to numerical probabilities. From this perspective, proponents of infinitesimal probabilities may claim four possible advantages over traditional numerical belief networks:

1. It may be easier to express beliefs by partitioning events into a small number of sets of relative plausibility, that is values, than by assigning each event a precise numerical probabilities.
2. Results from reasoning with infinitesimal probabilities are more robust and therefore more trustworthy since they are based on less specific inputs.
3. Reasoning with infinitesimal probabilities is easier to understand and explain.
4. Inference methods with infinitesimal probabilities can be computationally more efficient.

Hitherto, these claims have been largely untested. Initial analysis of the computational complexity of reasoning infinitesimal probabilities Darwiche [1992] suggests that, in general, it is of the same order as reasoning with numerical probabilities, that is NP-hard [Cooper, 1990]. There may be modest computational savings from doing arithmetic with small integers instead of floating point numbers.

Most research on qualitative probabilities has concentrated on developing the formalisms and efficient algorithms. There has been little concerted effort to demonstrate their application to real tasks and to evaluate their practicality. Initial studies of QPNs [Henrion and Druzdzel, 1990; Druzdzel and Henrion, 1993; Druzdzel, 1993] suggest that they are often inconclusive for nontrivial cases. For example, QPNs give vacuous results in any case with conflicting evidence. Studies of qualitative simulation have found similar difficulties. Much current research on qualitative simulation is directed towards integrating quantitative information to resolve ambiguities (and the resultant combinatorial explosions of the search space).

In this paper, we report the results of an initial experimental study comparing the diagnostic performance on a specific belief network using (1) the κ-calculus or infinitesimal probabilities, and (2) numerical probabilities. Our goal is to examine how well the infinitesimal scheme performs as an approximation to the numerical representation. We start with a fully assessed numerical representation, convert this into a kappa-representation using finite ε values, and perform inference on a set of test cases. We first explain the mappings we used to obtain infinitesimal or κ-values from the numerical probabilities, and how we mapped back from the posterior κ-values into probabilities for comparison of performance. Then, we describe the experimental design, including the sample network, the set of test cases, and our variations of the prior fault probabilities, the epsilon values used in mapping, and the number of findings observations per case. The infinitesimal scheme provides a set of the most plausible diagnoses for each case. In the results, we compare these plausible sets with the posterior probabilities for the diagnoses produced by the numerical scheme. Finally, we discuss the implications of these results for the application of the κ-calculus as a practical representation.

## 2 MAPPINGS BETWEEN NUMERICAL AND INFINITESIMAL PROBABILITIES

In order to be able to apply the κ–calculus to probabilistic reasoning on a belief network with finite probabilities, we need to provide a mapping from probabilities into kappa values. In order to compare the results we need to map the kappa results back again into probabilities. Strictly, the κ–calculus is only valid as ε→0. We use an approximation for finite values of ε. For a finite ε, the κ–calculus partitions the real interval [0,1] into regions identified by integers, based on the smallest power of in the polynomial. This mapping is illustrated in Figure 1.

More specifically, consider the real [0,1] interval $I$, which is the interval used by probability theory, and a discretized representation of $I$, which we call $S$. $S$ is a set of non-negative integers which the -calculus uses to represent probability measures in the interval $I$. We wish to explore the mappings $f: I \to S$ (i.e., from numerical to infinitesimal probability) and $g: S \to I$ (i.e., from infinitesimal to numerical probability). Note that there is information loss in the mapping $f$, since it is not injective. Moreover, the mapping $g$ is not surjective.

**Definition 1** [$\kappa$-map] [Spohn 1988] The mapping $f$ from probability measures to $\kappa$-values takes a probability $\pi$ and a threshold probability $\varepsilon$ and outputs a $\kappa$-value $\kappa \in S$ such that

$$f(\pi) = \kappa \quad \text{if} \quad \varepsilon^\kappa > \pi \geq \varepsilon^{\kappa+1}. \quad (1)$$

Figure 1 shows an example of a mapping for $\varepsilon = 0.1$.

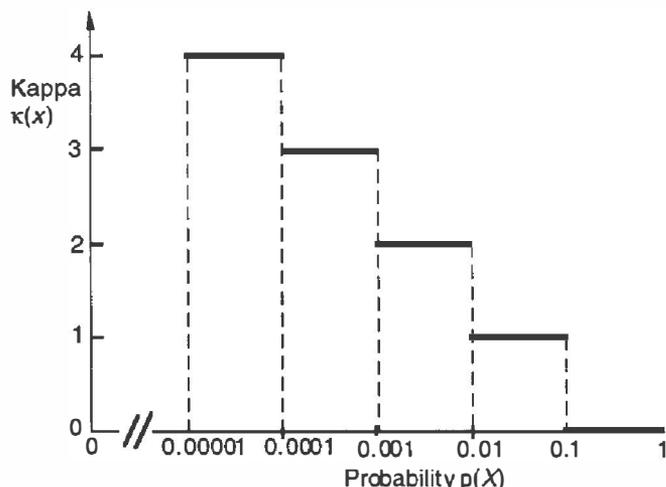

**Figure 1**: An example mapping giving kappa as a function of probability, for $\varepsilon=0.1$.

## 3 APPLICATION DOMAIN: WHY YOUR CAR DOES NOT START

The task is to troubleshoot why a car is not starting, given evidence on the status of the lights, battery, fuel, fan belt, and so on. Figure 2 shows the Bayesian belief network displaying the causal and conditional independence relations. We are grateful to David Heckerman for providing the original belief network and to Paul Dagum for lending us his expertise as a car mechanic in adjusting some of the probabilities. All variables are binary (present or absent), except for battery charge which has three values (high, low, none). The initial network contains fully quantified, numerical conditional probability distributions for each influence and prior probabilities for each fault (source variable). Effects of multiple causes of a common effect are combined with noisy-ORs, generalized where necessary.

There are nine explicitly identified faults in this model:
    spark plugs bad
    distributor bad
    fuel line bad
    fuel pump bad
    gas tank empty
    starter bad
    battery bad
    fan belt loose
    alternator bad

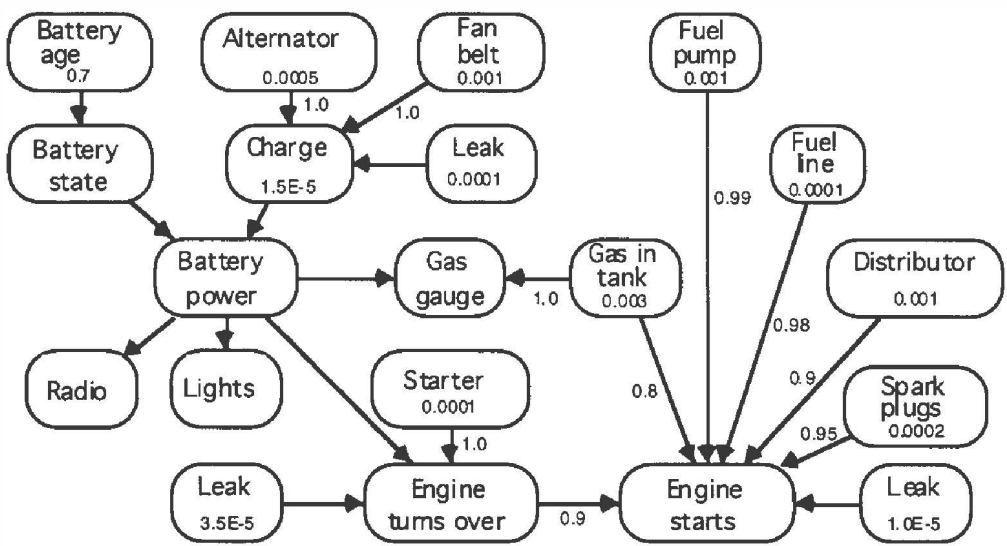

**Figure 2**: Bayesian network representing the car diagnosis domain. Leak events represent all the potential causes of a fault other than those shown explicitly. The number in each origin fault of a leak node represents its prior probability in the original network. The numbers attached to each influence arrow represent causal strengths — that is the probability that the successor is broken given that the predecessor is broken, and all other predecessors are normal.

We also identified three leaks. Each leak event represents all possible causes of an event that are not explicitly identified above. The probability of a leak is the probability that its associated effect will be observed even though none of its identified causes are present.

    engine start other
    engine turn over other
    charging system other

The *leaky noisy or* model assigns a probability to each leak, to handle the fact that the network is inevitably incomplete. In our adjusted network, the probability of each leak was substantially smaller than the sum of the probabilities of the identified causes for each event.

There are 10 observable findings in the model, listed here in non-decreasing order of expense to test:
1. engine-start
2. gas-gauge
3. engine-turn-over
4. lights
5. radio
6. fan-belt
7. battery-age
8. distributor
9. spark-plugs
10. alternator

Note that there are four findings that are also enumerated faults, namely fan belt, alternator, spark plugs, and distributor.

## 4 EXPERIMENTAL DESIGN

We wish to investigate the effects of three factors on the diagnostic performance of infinitesimal probabilities:

(a) The choice of the value of $\varepsilon$ on the mapping between numerical and infinitesimal probabilities.

(b) The range of prior fault probabilities

(c) The quantity of evidence in the test cases.

We have already discussed factor (a). Here, we will discuss our choice of each of these factors, and the conduct of the experiment.

### 4.1 Range of prior fault probabilities

The numbers in Figure 2 are the original prior fault probabilities. To examine the effect of the magnitude of the priors on the relative performance of the infinitesimal calculus, we created versions of the network with larger probabilities. To do this, we multiplied the prior odds by an odds factor ranging from 10 to 1000. Table 1 shows the mean and range of the resulting prior odds we used.

**Table 1:** The minimum, mean, and maximum prior fault probabilities. The top line shows the original probabilities. Those below are derived by multiplying the odds of each prior by the odds factor and converting back to probabilities.

| Odds factor | Minimum | Mean | Maximum |
|---|---|---|---|
| 1 | 0.00001 | 0.00036 | 0.00100 |
| 10 | 0.00010 | 0.00361 | 0.00991 |
| 50 | 0.00051 | 0.01750 | 0.04766 |
| 100 | 0.00103 | 0.03376 | 0.09099 |
| 300 | 0.00307 | 0.08900 | 0.23095 |
| 1000 | 0.01017 | 0.21364 | 0.50025 |

### 4.2 Test Cases and quantity of evidence

We expected that the performance of both numerical and infinitesimal schemes would improve as a function of the quantity of evidence. We also wished to examine the effect of the quantity of evidence on the relative performance of the two schemes. Accordingly, we needed a representative set of test cases with varying numbers of findings.

We generated a set of 116 test cases, in the following manner: For each of twelve faults (nine identified faults plus three leaks), we identified the most likely (modal) value for each of the ten observable findings. For each fault, we created a base case consisting of all findings at their modal value. In four cases, the fault is itself a finding, which we omitted from the base test case, since including the true fault as observed in the test case would be trivial. We then generated a second case for each fault by omitting the most expensive observation from the base case. Further cases were generated by omitting the next most expensive observation in turn. In all cases, we retained the finding that the engine does not start. In this way, we created a series of ten cases for eight faults, and nine cases for the four faults that are observable, resulting in a total of 116 test cases in all.

## 4.3 Computation

To obtain results for the numerical probabilistic scheme, we employed IDEAL [Srinivas and Breese, 1990], using the clustering algorithm from the IDEAL library. We applied each of the 116 test cases to the network using each of the six sets of priors, performing a total of 696 run. For each run we computed the posterior probability for each of the twelve faults resulting in 8352 probabilities.

We also converted the original numerical probabilities into $\kappa$-values, using the three values $\varepsilon$ (0.1, 0.01, 0.001), resulting in a total of 2088 additional runs. We ran each case using CNETS, a full implementation of the $\kappa$-calculus developed at the Rockwell Palo Alto Laboratory [Darwiche, 1994], producing posterior $\kappa$-values for each fault. For each run, we computed the *plausible set*, that is the subset of faults with the minimal $\kappa$ value.

**Definition 2** [Plausible Set] Consider a set $V = \{v_1, v_2, ..., v_m\}$ representing $m$ possible hypotheses, in which each hypothesis has been assigned a $\kappa$-value. Let $v_{\min} = \min_j v_j$ by the minimum $\kappa$-value.
The plausible set is given by

$$\Phi(V) = \{j : v_j = v_{\min}\}.$$

To compare the infinitesimal scheme with the numerical one, we converted $\kappa$-values of diagnoses back to probabilities as follows:

**Definition 3**: [Probability score] For a set $V = \{v_1, v_2, ..., v_m\}$ representing $m$ possible hypotheses, in which each hypothesis has been assigned a $\kappa$-value, the corresponding probability distribution is given by

$$\pi_j = \begin{cases} \dfrac{1}{|\Phi(V)|} & \text{if } v_j = v_{\max} \\ 0 & \text{otherwise} \end{cases} \quad (3)$$

That is, the probability $\pi_j = 1/n$ is assigned to the true faults if it is in the plausible set of size $n$. Otherwise, we assigned $p = 0$.

As an additional test, we also ran IDEAL using the exact algorithm, but using fault probabilities mapped to $0.01^\kappa$ for the values obtained from the mapping using the full set of $\kappa$ values. We applied this to a subset of 72 test cases. In the results, the plausible faults are clearly identifiable, having probabilities at least an order of magnitude greater than those of all other faults. We found that this approach, as expected, gave very similar results to the exact $\kappa$-calculus inference using CNETS.

# 5 RESULTS

Our first goal was to examine the effect of $\varepsilon$ values on the performance of the infinitesimal probability scheme. We then selected the value of $\varepsilon$ that gave the best results and examined the effect of varying the quantity of evidence on the performance of both numerical and infinitesimal schemes.

## 5.1 Effect of $\varepsilon$ values

Since the kappa calculus is only strictly correct as $\varepsilon \to 0$, we might expect it to perform better for small $\varepsilon$, where the approximation will be more exact. On the other hand, a larger $\varepsilon$ provides more partitions to the probability interval (0, 1], as shown in Figure 1, and consequently, it provides a finer discretization of the original probabilities, with less information lost. Accordingly, we might expect it to do better with larger $\varepsilon$. To investigate this we analyzed an initial set of 72 test cases using $\varepsilon$ values of 0.0001, 0.001, 0.01, 0.1, 0.2. Figure 3 shows a graph of average probability score assigned to the true diagnosis for these cases, against $\varepsilon$. It is interested to note that the average score is identical for $\varepsilon = 0.01$ and $\varepsilon = 0.001$, and also identical for $\varepsilon = 0.1$ and $\varepsilon = 0.2$. Overall, there is an improvement in performance with increasing $\varepsilon$ up to 0.2. Accordingly, we selected $\varepsilon = 0.1$ for use in our remaining experiments.

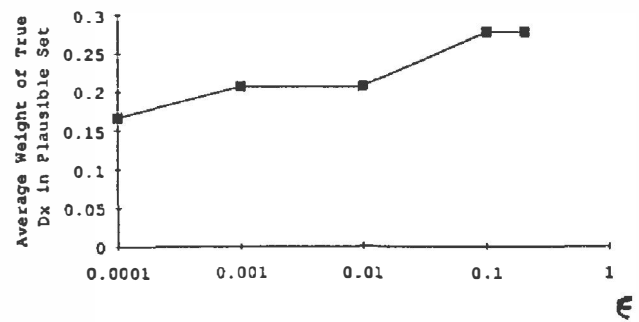

Figure 3: Effect of $\varepsilon$ on the score (probability assigned to the true fault) by the infinitesimal scheme

## 5.2 Effect of Number of Findings on the Plausible set

As the quantity of evidence increases, we should expect the performance of both numerical and infinitesimal schemes to improve. Accordingly, we classified the cases by the number of findings. Figure 4 graphs the average size of the plausible set (number of plausible faults) identified by the infinitesimal scheme as a function of the number of findings. These results summarize all 116 cases for $\varepsilon = 0.1$. As expected, the average size of the plausible set of faults decreases with the number of findings, from 7 faults with 1 finding to 1.21 faults for 10 findings. With 10 findings, this scheme provides almost complete specificity that is, the plausible set usually consists of just a single diagnosis.

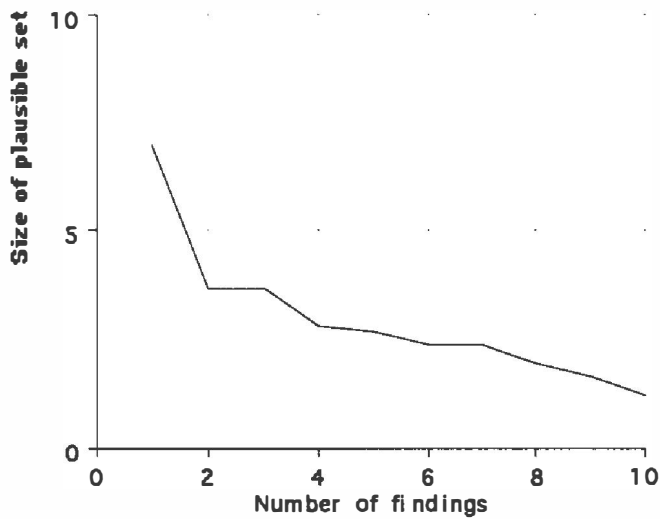

**Figure 4:** The average size of the plausible set as a function of the number of findings in each case.

## 5.3 Comparing the performance of infinitesimal and numerical schemes

Next, we compare how the number of findings affects the diagnostic performance for the infinitesimal and numerical schemes. Figure 5 graphs the performance in terms of the average probability each assigns to the true fault, as a function of the number of findings. For both schemes, as expected, the average probability assigned to the true fault increases with increasing evidence, from about 0.15 with 1 finding, to about 0.47 with 10 findings.

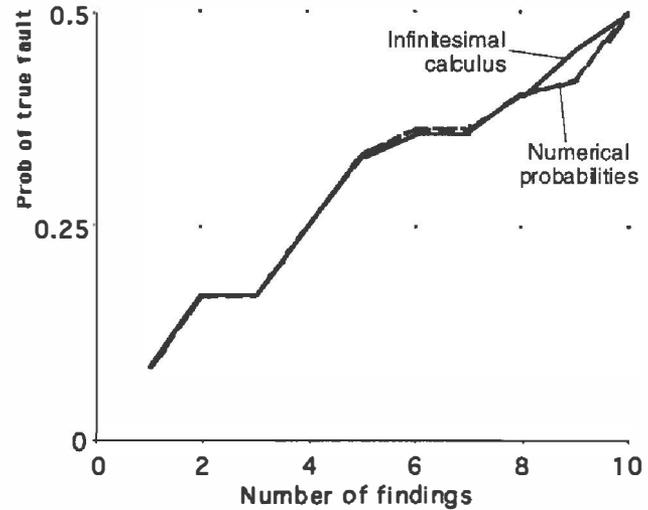

Figure 5: The probability assigned to the true fault for each scheme as a function of number of findings

What is, perhaps, surprising is how closely the performance of the infinitesimal scheme tracks the performance of the numerical scheme. Indeed the infinitesimal scheme appears to perform better than the numerical scheme for intermediate numbers of findings, but this difference is not significant. Since the infinitesimal representation is derived from the numerical one, we could not expect it to do better, on average.

Note that, even with all ten findings, both schemes average about 0.5 probability for the true diagnosis. This relatively poor performance arises because of the limited scope of the network, which does not provide the means to differentiate among several classes of fault.

## 5.3 The magnitude of priors and the performance of infinitesimal probabilities

The infinitesimal probability scheme appears to perform very well relative to numerical probabilities for the original car network, in which the prior fault probabilities are very small, on average 0.00036 To examine if it performs equally well for larger priors, we multiplied the prior odds by five odds factors, as shown in Table 1. Figure 6 shows the average probability assigned to the true diagnosis as a function of the average priors. Interestingly, the two schemes are almost indistinguishable up to an average fault prior 0.033. Above that, the performance of the infinitesimal probability drops off sharply — that is, for average priors of 0.089 and 0.214. These results confirm our expectation that infinitesimal works well for small priors, but not so well for large priors.

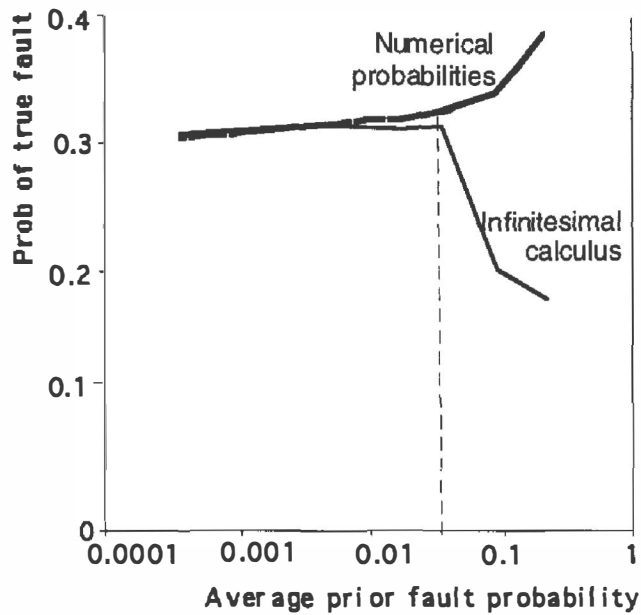

**Figure 6:** Comparison of the average performance of infinitesimal and numerical probability schemes as a function of prior fault probabilities.

## 6 CONCLUSIONS

We find these initial results very encouraging in terms of the diagnostic performance of the infinitesimal probability scheme. For this example domain, we found the best performance occurs using $\varepsilon = 0.1$ to $0.2$. Performance for $\varepsilon = 0.01$ was slightly worse. Performance of the infinitesimal scheme relative to the numerical scheme does not appear to vary significantly with the quantity of evidence. The performance using infinitesimal probability is not noticeably worse than the numerical probabilities for prior fault probabilities up to about 0.03. For larger average fault probabilities, the relative performance of infinitesimal probabilities starts to drop off sharply. This findings suggests that infinitesimal probabilities are more likely to be reliable for diagnosis tasks with very small prior fault probabilities, such as most machine and electronic devices. They may also work for some medical domains, as long as the disease priors are less than 1%.

The mapping from $\kappa$-values back to probabilities that we have used is very simple. More sophisticated mappings are possible, making use of higher values. We should also point out that the scoring methods that we have used to evaluate performance have been based on posterior probability of the true diagnosis, which is perhaps the most exacting way to compare them. Another way would be to compare the rank ordering of diagnosis. A third, would be to evaluate the quality of decisions based on the diagnosis. In general, scoring rules based on ranks of diagnosis or, even more, the quality of decisions will be less rather than more sensitive to these differences in representation.

While these findings are encouraging for the practical usefulness of infinitesimal probabilities, we should remember that these initial results are on a single domain. This car model domain is simple, with few loops and short chains. This kind of experiment should be conducted on a wide range of types of network to see how far these initial results will hold up.

In the introduction, we distinguished view (a) of infinitesimal probabilities, as an approach to nonmonotonic reasoning, from view (b), as an approximation to numerical probabilities. We reiterate that this paper, we focus on (b), and we are not attempting to evaluate its use as an approach to nonmonotonic logic. Conclusions about the former have limited relevance to the latter.

Infinitesimal probabilities are quite appealing as an alternative to numerical probabilities. They should be significantly easier to elicit from experts. Inference may be more efficient. And resulting inferences should be somewhat more robust to changes in probabilities.

Some questions that need further investigation include:

> Does the best choice of $\varepsilon$ vary with the domain?
>
> Does these results hold for larger networks, with more complex structures?
>
> Can this infinitesimal approximation be extended to utilities and decision making?
>
> Can we obtain a clearer analytic characterization of when performance will be or won't be reliable?

In addition, we need practical knowledge engineering methods for eliciting infinitesimal probabilities. We anticipate that, in the long run, the best practical tools will combine qualitative and quantitative methods.

## Acknowledgments

This work was supported by the National Science Foundation under Grant Project IRI-9120330 to the Institute for Decision Systems Research. We would like to thank David Heckerman for use of the car diagnosis network, and Paul Dagum for help in refining some of the probabilities.